\documentclass[10pt,twocolumn,letterpaper]{article}
\usepackage{float}
\usepackage{iccv}
\usepackage[linesnumbered,ruled]{algorithm2e}
\usepackage{times}
\usepackage{makecell}
\usepackage{epsfig}
\usepackage{amsmath}
\usepackage{arydshln}
\usepackage{amsthm}
\usepackage{amssymb}
\usepackage{cite}
\usepackage{graphicx} 
\usepackage{subfigure}	
\usepackage[table]{xcolor}
\usepackage{diagbox}
\usepackage{multirow}
\usepackage{color}
\usepackage{booktabs}
\usepackage{pifont}
\newtheorem{Proposition}{Observation}

\usepackage[pagebackref=true,breaklinks=true,colorlinks,bookmarks=false]{hyperref}

\iccvfinalcopy 
\let\titleold\title
\renewcommand{\title}[1]{\titleold{#1}\newcommand{\thetitle}{#1}}
\def\maketitlesupplementary
   {
   \newpage
       \twocolumn[
        \centering
        \Large
        \textbf{\thetitle}\\
        \vspace{0.5em}Supplementary Material \\
        \vspace{1.0em}
       ] 
   }

\def\iccvPaperID{2672} 

\ificcvfinal\fi

\begin{document}

\title{High-Fidelity Differential-information Driven Binary Vision Transformer}

\author{
Tian Gao$^{1}$\hspace{0.02in}
Zhiyuan Zhang$^{2}$ \hspace{0.02in}
Kaijie Yin$^{1}$ \hspace{0.02in}
Chengzhong Xu$^{1}$ \hspace{0.02in}
Hui Kong$^{1,}$ \thanks{Corresponding author}
\vspace{0.1in}\\
$^{1}$University of Macau \quad
$^{2}$Singapore Management University
}

\maketitle
\ificcvfinal\thispagestyle{empty}\fi
\begin{abstract}
The binarization of vision transformers (ViTs) offers a promising approach to addressing the trade-off between high computational/storage demands and the constraints of edge-device deployment. However, existing binary ViT methods often suffer from severe performance degradation or rely heavily on full-precision modules. To address these issues, we propose DIDB-ViT, a novel binary ViT that is highly informative while maintaining the original ViT architecture and computational efficiency. Specifically, we design an informative attention module incorporating differential information to mitigate information loss caused by binarization and enhance high-frequency retention. To preserve the fidelity of the similarity calculations between binary $Q$ and $K$ tensors, we apply frequency decomposition using the discrete Haar wavelet and integrate similarities across different frequencies.  Additionally, we introduce an improved RPReLU activation function to restructure the activation distribution, expanding the model’s representational capacity. Experimental results demonstrate that our DIDB-ViT significantly outperforms state-of-the-art network quantization methods in multiple ViT architectures, achieving superior image classification and segmentation performance.
\end{abstract}

\section{Introduction}
Transformer has advanced natural language processing (NLP) thanks to the ability to encode long-range dependencies. Inspired by the remarkable success, vision transformers (ViT) have also been extensively investigated and applied to various vision tasks, including object detection~\cite{li2022exploring,zhang2021vit}, classification~\cite{touvron2021training,liu2021swin}, and image segmentation~\cite{strudel2021segmenter,yang2022lavt}. Despite effectiveness, the large parameter size and high memory demands of transformer models preclude the deployment of resource-constrained devices. Addressing this limitation has become imperative, making ViT compression an active area of research. In recent years, several model compression techniques have been proposed, including Network Pruning~\cite{gao2021network}, Low-rank Approximation~\cite{markovsky2012low}, low-bit quantization~\cite{zhuang2018towards,li2022q} and model binarization~\cite{liu2020reactnet,chen2021bnn,rastegari2016xnor,lin2022siman}, among which model binarization has attracted significant attention for achieving higher compression ratio, making them particularly promising for practical applications~\cite{qin2020binary,qin2023bibench}.  

Over the past decade, significant efforts have focused on binarizing Convolutional Neural Networks (CNNs) to alleviate computational and memory bottlenecks. For example, Xnor-Net~\cite{rastegari2016xnor} introduces scale factors to enhance the performance of Binary Neural Networks (BNNs)\cite{hubara2016binarized} on the ImageNet-1k dataset~\cite{deng2009imagenet}. ReAct-Net\cite{liu2020reactnet} further advances this area by achieving remarkable results with low complexity while maintaining full-precision performance on large datasets, such as ImageNet-1k. Additionally, FDA~\cite{xu2021learning} leverages Fourier transform results with frequency truncation to estimate the gradient in the binarization process. PokeBNN~\cite{zhang2022pokebnn} improves the quality of BNNs by adding multiple residual paths and tuning the activation function with higher accuracy.  To further enhance accuracy, BNext~\cite{guotowards} increases the model’s parameter size, achieving over 80\% accuracy on ImageNet-1k.

Despite their effectiveness, CNN binarization methods are not directly applicable to ViTs due to the unique characteristics of the attention module. To bridge the gap between CNN and ViT binarizations, BiViT~\cite{he2023bivit} is the first to customize the ViT structure, proposing Softmax-aware Binarization to reduce binarization errors. This approach, however, partially relaxes binary constraints, leaving MLP activations in full precision. BinaryViT~\cite {le2023binaryvit} transfers additional operations (e.g. multiple average pooling layers~\cite{le2023binaryvit}) from the CNN architecture into a binary pyramid ViT architecture. To further improve the binarization quality, subsequent work Bi-ViT~\cite{li2024bi} proposes a channel-wise learnable scaling factor to mitigate attention distortion, BVT-IMA~\cite{wang2024bvt} utilizes look-up information tables to enhance model performance, GSB~\cite{gao2024gsb} designs group superposition binarization to improve the performance of the binary attention module. SI-BiViT~\cite{yin2024si} incorporated spatial interaction in the binarization process to enhance token-level correlations. Although these algorithms have unique merit, the performance of binary ViTs remains significantly below that of their full-precision counterparts due to the information loss during binarization. Moreover, these improvements often rely on maintaining full-precision components or computational modules, which adds complexity and undermines the efficiency of the binary models.

In this study, we aim to narrow the performance gap between binary ViTs and their full-precision counterparts while preserving the original ViT architecture. To achieve this, we propose DIDB-ViT, a highly informative binary ViT designed to minimize the disparity with its full-precision counterpart without increasing computational complexity. Specifically, we conduct an in-depth analysis of the computational differences between binarization and full-precision operations across the entire attention mechanism. Based on these findings, we propose a series of targeted strategies for various ViT components to maximize the informativeness of the binary ViT. In particular, we design an informative attention module employing differential information to mitigate the information loss caused by binarization and enhance the retention of high-frequency information. To preserve the fidelity of similarity calculations in binary $Q$ and $K$, we optimize the process by decomposing the input to obtain high- and low-frequency components. Then, the final faithful similarity is obtained by integrating the similarity between binary Q and K across each frequency. Furthermore, we propose an improved  RPReLU to restructure the overall distribution of activations to further increase the representational space of the model's outputs.

Our contributions are summarized as follows:\\ 
\indent $\bullet$ We design an \textbf{informative attention module} that incorporates differential information to address the information loss caused by binarization and enhance high-frequency information retention, improving the overall effectiveness of binary ViTs. \\
\indent $\bullet$ We propose \textbf{frequency-based binarization} for the $Q$ and $K$. Based on a discrete Haar wavelet and four binary linear layers, the input is converted into the $Q$ and $K$ tensors based on high and low-frequency components. The high-fidelity similarity is obtained by integrating the similarity between binary $Q$ and $K$ across different frequencies. \\
\indent $\bullet$ We propose an \textbf{improved RPReLU} activation function to restructure the overall distribution of activations, expanding the model’s representational capacity and improving its performance without increasing computational complexity. \\
\indent $\bullet$ The experimental results demonstrate that our approach achieves \textbf{SOTA performance} among the existing binary vision transformer in both small and large datasets, such as CIFAR-100, TinyImageNet, and ImageNet-1K.

\section{Related work}
\noindent\textbf{CNN Binarization.}~Early studies on neural network binarization focus on Convolutional Neural Networks (CNNs). Foundational works like BinaryConnect~\cite{courbariaux2015binaryconnect} introduce binary weight networks, employing binary weights while retaining full-precision activations. Binarized Neural Networks (BNNs)~\cite{hubara2016binarized} extends binarization to both weights and activations, marking a significant milestone in binary deep learning. Subsequently, XNOR-Net~\cite{rastegari2016xnor} introduces scale factors to enhance the representational capability of binary models, achieving promising results on ImageNet-1K~\cite{deng2009imagenet}. ABC-Net~\cite{lin2017towards} alleviated information loss by using multiple binary weights and multiple binary activations. 
To further improve the performance of binary CNNs, advanced techniques such as ReActNet~\cite{liu2020reactnet} and Bi-Real Net~\cite{liu2018bi} refine activation functions and incorporate residual structures, reducing the performance gap between binary and full-precision models. Similarly, FDA~\cite{xu2021learning} leverages frequency domain analysis to provide more accurate gradient estimates during binarization, resulting in notable accuracy gains. IR-Net~\cite{qin2020forward} reduces the information loss of both weights and activations without additional operation on activations, while PokeBNN~\cite{zhang2022pokebnn} further enhances the quality of binary networks by using multiple residual paths and tuning the activation function. ReBNN~\cite{xu2023resilient} reduces binarization error by parameterizing the scaling factor and a weighted reconstruction loss. More recently, BNext~\cite{guotowards} increases model capacity with parameter scaling, achieving over 80\% accuracy on ImageNet-1K. Despite these advancements, applying CNN binarization techniques directly to ViT models remains challenging, leading to increased research interest in ViT binarization.
\\
\textbf{ViT Binarization.}~ BiViT~\cite{he2023bivit} is the pioneering work on ViT Binarization which proposes a Softmax-aware Binarization to handle the long-tailed
 distribution of softmax attention. Subsequently, Li et al.~\cite{li2024bi} analyze the causes of performance degradation in ViT binarization and propose a channel-wise learnable scaling factor to mitigate attention distortion, and ranking-aware distillation is also employed to enhance model training. BVT-IMA~\cite{wang2024bvt} reveals the quantity of information hidden in the attention maps of binary ViT and proposes a straightforward approach to enhance model performance by modifying attention values using look-up information tables. To deal with the information loss, GSB~\cite{gao2024gsb} introduces group superposition binarization to improve the performance of the binary attention module. SI-BiViT~\cite{yin2024si} incorporates spatial interaction in the binarized ViTs by using a plug-and-play dual branch to enhance the token-level correlation with improved performance on classification and segmentation. BFD~\cite{li2024bfd} applies frequency-enhanced Distillation to improve the performance of binary ViT. Despite these advancements, higher accuracy is often achieved by introducing additional computational complexity (e.g. full-precision module). This motivates our proposal of a fully binary ViT method that maintains high accuracy without relying on overly large full-precision components.
\section{Method}
\subsection{Preliminaries}


In binary models, weights and activations are constrained to {-1,1} or {0,1}, enabling efficient multiplications using $Xnor$ and $popcount$ operations.
Let's take the binarization of the linear layer ($\mathrm{BL}()$), a core component of ViTs for channel-wise feature aggregation, as an example,
\vspace{-5pt}
\begin{equation}
\begin{aligned}
\label{blinear}
\mathbf{O} =\mathrm{BL}\left( \mathbf{A},\mathbf{W} \right)= \boldsymbol{\alpha}\odot \left( \hat{\mathbf{A}} \otimes \hat{\mathbf{W}} \right) ,
\end{aligned}
\vspace{-5pt}
\end{equation}
where $\mathbf{O}$, $\mathbf{A}$, $\mathbf{W}$, $\hat{\mathbf{W}}$, and $\hat{\mathbf{A}}$ represent the output, activation, weight, binarized weight, and binarized activation, respectively. $\boldsymbol{\alpha}$ is a channel-wise scaling factor. $\odot$ and $\otimes$ denote element-wise multiplication and binary matrix multiplication, respectively.
The binarized activation $\hat{\mathbf{A}}$  is computed using a $sign$ function for forward propagation and a piecewise polynomial function for the backward pass:
\begin{equation}
\label{eq2-1}
\begin{aligned}
&\mathbf{Forward}~\hat{\mathbf{A}}=\mathrm{B}\left( \mathbf{A},a,b \right) =sign\left( \frac{\mathbf{A}-b}{a} \right) ,\\
	&\mathbf{Backward}~\frac{\partial L}{\partial \mathbf{A}}=\frac{\partial L}{\partial \hat{\mathbf{A}}}\odot\frac{\partial \hat{\mathbf{A}}}{\partial \mathbf{A}}=\\
 &\left\{ \begin{matrix}
	\frac{\partial L}{\partial \hat{\mathbf{A}}}\odot\left( 2+2\left( \frac{\mathbf{A}-b}{a} \right) \right)&		b-a \leqslant \mathbf{A}<b\\
	\frac{\partial L}{\partial \hat{\mathbf{A}}}\odot\left( 2-2\left( \frac{\mathbf{A}-b}{a} \right) \right)&		b\leqslant \mathbf{A}<b+a\\
	0&		otherwise\\
\end{matrix} \right. ,\\
\end{aligned}
\end{equation}
where $a$ and $b$ are learnable scaling and bias terms, respectively, and $L$ is the loss function. For attention values (ranging from 0 to 1), a specific binarization method is applied:
\begin{equation}
\label{eqatt}
\begin{aligned}
	\mathbf{Forward}&~\hat{\mathbf{A}}_{att}=\mathrm{B}_{att}\left( \mathbf{A}_{att},a,b \right)\! =\\
 &a \times clip\left( round\left( \frac{\mathbf{A}_{att}-b}{a} \right) ,0,1 \right),\\
	\mathbf{Backward}&~\frac{\partial L}{\partial \mathbf{A}_{att}}=\left\{ \begin{matrix}
	a\times\frac{\partial L}{\partial \hat{\mathbf{A}}_{att}}&		b\leqslant \mathbf{A}_{att}<a +b\\
	0&		otherwise\\
\end{matrix} \right. ,
\end{aligned}
\end{equation}
where $\mathrm{B}_{att}$ binarizes the full-precision attention $\mathbf{A}_{att}$ to binary attention $\hat{\mathbf{A}}_{att}$. $clip\left( x,0,1 \right)$ ensures outputs remain in [0,1], and $round$ maps inputs to the nearest integers.

Weights are commonly binarized as follows:
\begin{equation}
\label{eqw}
\begin{aligned}
\mathbf{Forward}~&\mathbf{\hat{W}}_{\left[ :,k \right]}=\mathrm{B}_w\left( \mathbf{W}_{\left[ :,k \right]} \right) \\
&=G\left( abs\left( \mathbf{W}_{\left[ :,k \right]} \right) \right)
\odot sign\left( \mathbf{W}_{\left[ :,k \right]} \right) ,
\\
\mathbf{Backward}~&\frac{\partial L}{\partial \mathbf{W}_{\left[ :,k \right]}}=G\left( abs\left( \mathbf{W}_{\left[ :,k \right]} \right) \right) \\
&\odot \frac{\partial L}{\partial \mathbf{\hat{W}}_{\left[ :,k \right]}}\odot \mathbf{1}_{-1<\mathbf{W}_{\left[ :,k \right]}<1},
\end{aligned}
\end{equation}
where $\mathbf{W}_{\left[ :,k \right]}$ refers to the weights in the $k$-th channel, $G\left( \right)$ computes the scaling factor, and $\mathbf{1}_{-1<\mathbf{W}_{\left[ :,k \right]}<1}$ denotes a mask tensor with the same size as $\mathbf{W}_{\left[ :,k \right]}$. If the element in $\mathbf{W}_{\left[:,k \right]}$ falls in the closed interval [-1,1], the corresponding element of the mask tensor is marked as one.

\subsection{Differential-Informative Binary Attention}
\begin{figure}
\centering
\includegraphics[width=3.2in]{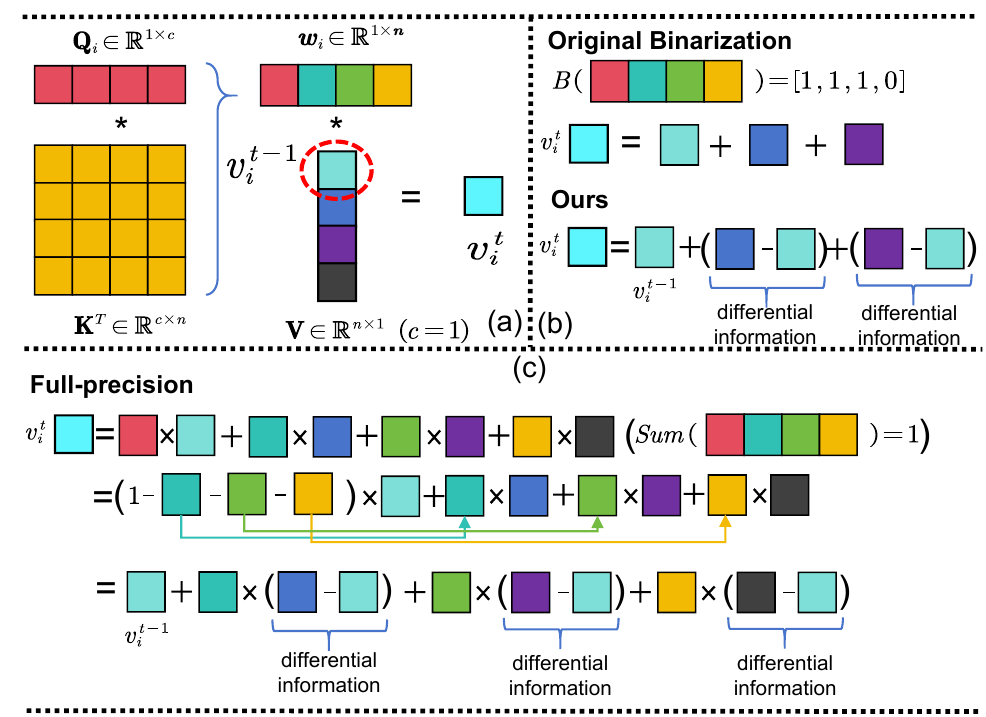}
\caption{The binary attention module based on differential information compensation. (a) Illustration of the attention module operation. (b) Comparison of binary feature updates between the original method and ours. (c) Illustration of the full-precision feature update process and the equivalent representation in differential form.}
\label{figdi}
\end{figure} 
Prior studies on binary ViT models~\cite{li2024bi,gao2024gsb} indicate that binarizing the attention matrix can result in significant performance degradation. To address this issue, analyzing the changes in module outputs before and after binarization is crucial.

Let $\bold{V}=\left[ \boldsymbol{v}_1,\boldsymbol{v}_2,\cdots ,\boldsymbol{v}_n \right]^{T} \in \mathbb{R} ^{n\times c}
$ be the full-precision value matrix with $n$ tokens and each token has $c$ channels. For simplification, we set $c$ to 1, reducing $\bold{V}$  to a column vector $\boldsymbol{v}=\left[v_1,v_2,\cdots ,v_n \right]^{T}\in \mathbb{R} ^{n \times 1}$.
Let $\boldsymbol{w}=\left[ w_1,w_2,\cdots ,w_n \right] \in \mathbb{R} ^{1 \times n}$ be the $i^{th}$ row of the full-precisian attention matrix $\bold{A}$, then the update of $i^{th}$ token in $\boldsymbol{v}$ is calculated by $v_i^t=\boldsymbol{w}\cdot\boldsymbol{v}^{t-1}$ (Eq.\ref{EQ5}), with $t$ and $t-1$  are identifiers used to distinguish before and after the attention operation, 
\begin{equation}
\begin{alignedat}{2}
v_{i}^{t}&=w_1v_{1}^{t-1}+w_2v_{2}^{t-1}+\cdots \cdots +w_nv_{n}^{t-1},
\label{EQ5}
\end{alignedat}
\end{equation}

Applying the binarization (Eq.~\ref{eqatt}) to $\boldsymbol{w}$, Eq.\ref{EQ5} turns into 
\begin{equation}
\label{modifyatt2}
\begin{alignedat}{2}
 v_{i}^{t}= v_{i}^{t-1} + \sum_{\stackrel{j\in \boldsymbol{\varUpsilon}}{j \ne i}} v_{j}^{t-1},
\end{alignedat}
\end{equation}
where $\boldsymbol{\varUpsilon}$ is the set of indexes of the elements whose values are 1 in the binarized $\boldsymbol{w}$. The number of elements in $\boldsymbol{\varUpsilon}$ is $k$ with $k \ll n$. Note that the binary weight of $v_{i}^{t-1}$ in Eq.\ref{modifyatt2} is 1 after binarization of the full-precision attention matrix because the corresponding $w_i$ is a diagonal element of the full-precision attention matrix, which means $w_i$ should be the value larger than the binary threshold ($b$ in Eq.~\ref{eqatt})

Alternatively, we can reformulate the Eq.~\ref{EQ5} as follow,
\begin{equation}
\begin{alignedat}{2}
v_{i}^{t}&=\left( 1-\sum_{\stackrel{j=1,}{j \ne i}}^n w_j \right) v_{i}^{t-1}+\sum_{\stackrel{j=1,}{j \ne i}}^n{w_jv_{j}^{t-1}},
\\
&=v_{i}^{t-1}+\sum_{\stackrel{j=1}{j \ne i}}^n{w_j(v_{j}^{t-1}-v_{i}^{t-1})}.
\label{EQ7}
\end{alignedat}
\end{equation}
where we observed that the update of $v_{i}^{t}$  contains the differential information between $v_i^{t-1}$ and $v_j^{t-1}$. The differential form is derived based on the fact that each row of the attention matrix sums up to 1.


Comparing Eq.\ref{EQ7} with Eq.~\ref{modifyatt2}, we can see that significant differential information is discarded and the importance of all tokens has become equal after binarization in Eq.~\ref{modifyatt2}. To solve this problem, alternatively, we can use the differential form in updating $v_{i}^{t}$ (Eq.\ref{EQ7}) and get the reformulation result (Eq.~\ref{modifyatt3}) if the binarization (Eq.~\ref{eqatt}) is applied to the $\boldsymbol{w}$




\begin{equation}
\label{modifyatt3}
\begin{aligned}
 v_{i}^{t}&= v_{i}^{t-1} + \sum_{\stackrel{j\in \boldsymbol{\varUpsilon}}{j \ne i}} (v_{j}^{t-1}-v_{i}^{t-1}), \\
 &= v_{i}^{t-1} + \sum_{j\in \boldsymbol{\varUpsilon}} (v_{j}^{t-1}-v_{i}^{t-1}).
\end{aligned}
\end{equation}

For binary vectors, differential features are equivalent to calculating the similarity between binary vectors. Adding differential features can partially restore the distinction of importance between tokens compared to directly adding features of other tokens to update each element of the $\bold{V}$. To simplify the calculation, Eq.~\ref{modifyatt3} could be converted into:
\begin{equation}
\label{modifyatt3-1}
\begin{aligned}
v_{i}^{t} &=\left( 1-k \right) v_{i}^{t-1}+\sum_{j\in \boldsymbol{\varUpsilon}}{v_{j}^{t-1}}.
\end{aligned}
\end{equation}

\begin{figure}
\centering
\includegraphics[width=3.3in]{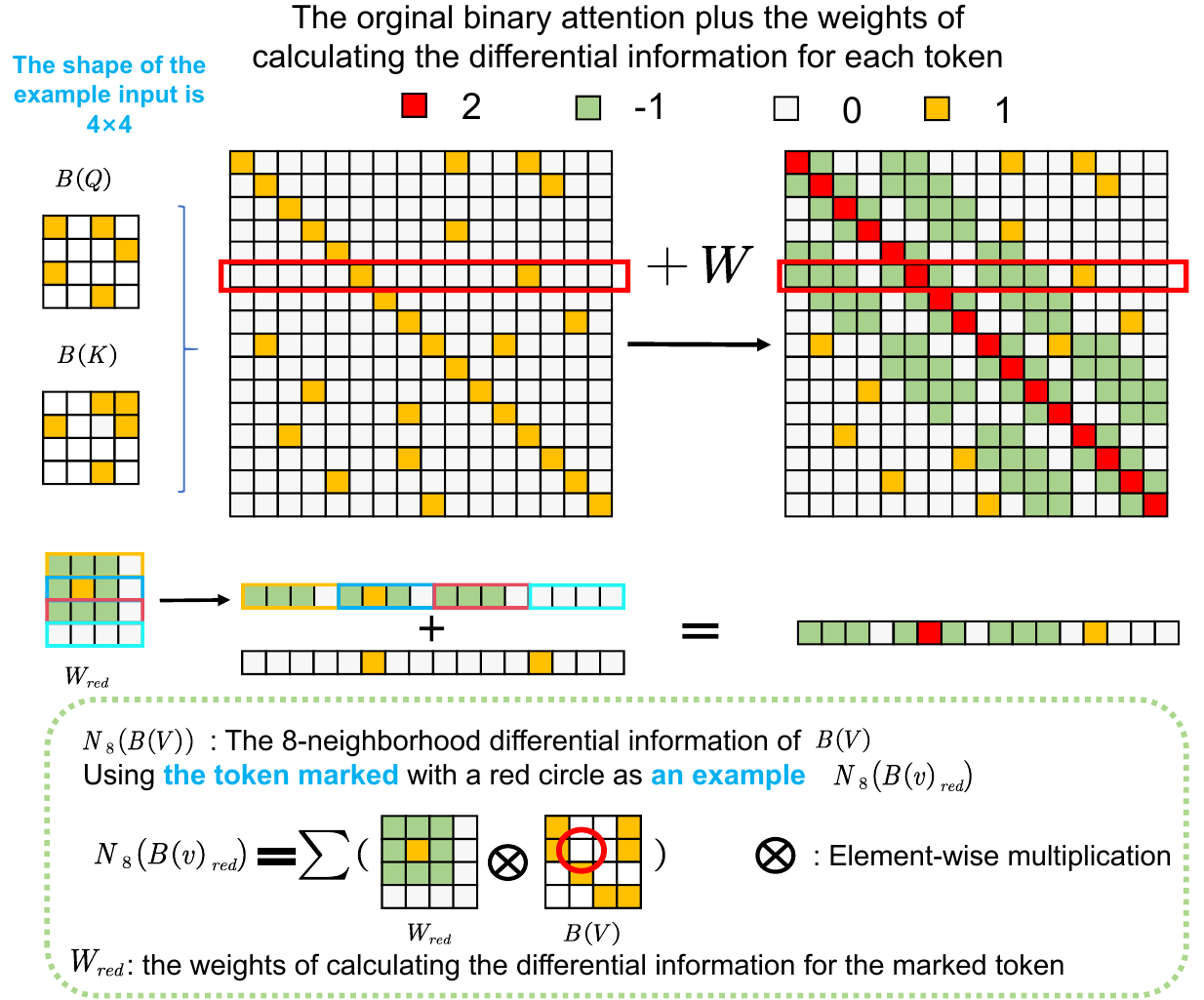}
\caption{The figure about mapping the eight neighborhoods of each token in the image-like arrangement to the corresponding positions in the attention matrix. The essence of $\sum_{l\in \varPsi}{\left( \mathrm{B}(v_{l}^{t-1}) \right)}$ is equivalent to setting -1 in the neighborhoods of each diagonal element in the attention matrix, which can be seen as negative attention information.}
\label{fignegatt}
\end{figure} 

Although Eq.~\ref{modifyatt3-1} achieves the partial retention of the differential information, the attention weights for each of the other tokens are still equal, reducing the roles of the different tokens when updating $\bold{V}$. Meanwhile, recent studies~\cite{bai2022improving,shin2024attentive} have shown that self-attention inherently behaves as a low-pass filter by calculating the weighted average of the value vectors associated with tokens, resulting in a substantial loss of high-frequency information. This issue is further exacerbated by the information loss introduced during binarization. 

\noindent\textbf{High-frequency Information Retention.} 
To retain high-frequency information and further restore the different importance of each token, we incorporate local differential information to Eq.~\ref{modifyatt5} as follows:
\begin{equation}
\label{modifyatt5}
\begin{aligned}
v_{i}^{t} &=\left( 1-k \right) v_{i}^{t-1}+ \sum_{j\in \boldsymbol{\varUpsilon}}{v_{j}^{t-1}}+\sum_{l\in \varPhi}{\left( v_{i}^{t-1}-v_{l}^{t-1} \right)},
\end{aligned}
\end{equation}
where $\varPhi$ represents the 8-neighborhood token features of $v_{i}$ (All tokens of $V$ are in an image-like arrangement.).  The additional introduced differential term enhances the high-frequency components of the features. Meanwhile, by utilizing the prior knowledge of Gaussian weight that the importance of the similarity between one token and other tokens is roughly inversely related to their spatial distance, it increases the value range of the binary attention matrix, and further partially restores the distinction of token importance based on similarity. 

To accelerate the proposed binarization processing, we decompose $\sum_{l\in \varPhi}{\left( v_{i}^{t-1}-v_{l}^{t-1} \right)}$ into $9 v_{i}^{t-1} $ and $-\sum_{l\in \varPsi}{\left( v_{l}^{t-1} \right)}$ ($\varPsi$ is a $3 \times 3$ receptive field). To further simplify, we set a learnable parameter $\beta$ as the coefficient of $v_{i}^{t-1}$ with an initial value of $ 10-k$. Then, we have
\begin{equation}
\label{modifyatt6}
\begin{aligned}
v_{i}^{t} &=\beta v_{i}^{t-1}+ \sum_{j\in \boldsymbol{\varUpsilon}}{v_{j}^{t-1}}-\sum_{l\in \varPsi}{v_{l}^{t-1}},
\end{aligned}
\end{equation}
\textbf{$\beta v_{i}^{t-1}$ can be easily implemented with a residual connection scaled by the factor $\beta$ without additional operation}. To obtain the final output of the binary attention module, we need to binarize each element of $\boldsymbol{v}^{t-1}$. To preserve more information, we retain $\beta v_{i}^{t-1}$ in full precision (shortcut), \textbf{avoiding any additional computational complexity}. Then, we have,
\begin{equation}
\label{modifyatt8}
\begin{aligned}
v_{i}^{t} &=\beta v_{i}^{t-1}+\alpha \sum_{j\in \boldsymbol{\varUpsilon}}{\mathrm{B}(v_{j}^{t-1})}-\gamma \sum_{l\in \varPsi}{\mathrm{B}(v_{l}^{t-1})},
\end{aligned}
\end{equation} 
where $\mathrm{B}()$ is the binary operator. As shown in Fig.~\ref{fignegatt}, the essence of $\sum_{l\in \varPsi}{\left( \mathrm{B}(v_{l}^{t-1}) \right)}$ is equivalent to setting -1 in the neighborhoods of each diagonal element in the attention matrix, which can be seen as negative attention information and efficiently implemented using a binary group convolution layer with frozen weights set to -1. and $\gamma$ is the scale factor. $\alpha \sum_{j\in \boldsymbol{\varUpsilon}}{\mathrm{B}(v_{j}^{t-1})}$ is the original output of the binary attention module with scale factor $\alpha$. 

\subsection{High-Fidelity Similarity Calculation}
\begin{figure}
\centering
\includegraphics[width=3.2in]{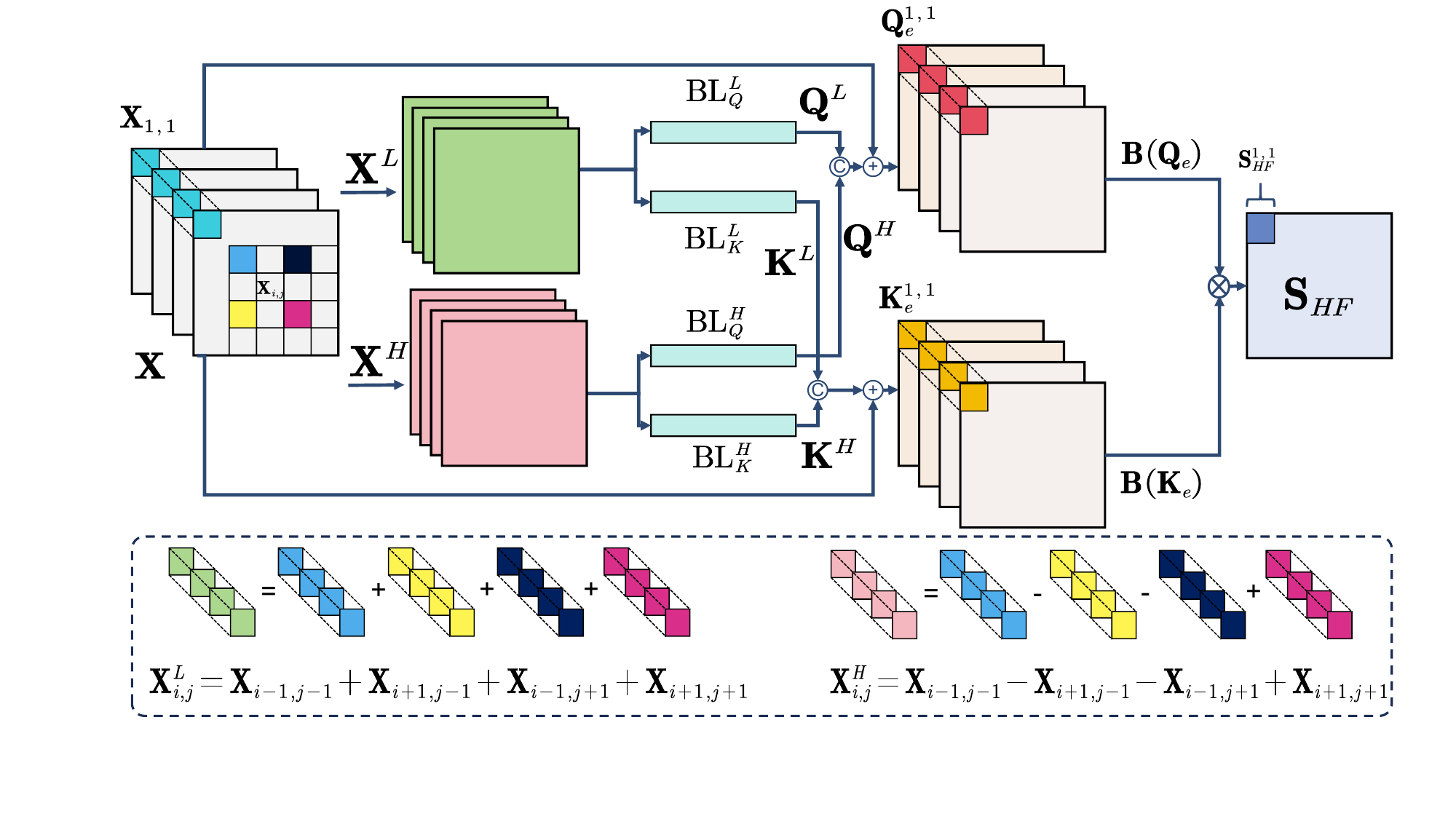}
\caption{The schematic diagram of modifying similarity map $\bold{QK}^T$ using non-subsampled discrete Haar wavelet decomposition. $\bold{X}_{i,j}$ is a element of input $\bold{X}$ in the position $(i,j)$. $\copyright$ is the concatenation operation. $\oplus$ means the element-wise summation and $\otimes$ indicate the dot product between the tokens of $\bold{Q}$ and the tokens of $\bold{K}$.}
\label{attentionmap}
\end{figure}
Binarization corrupts the similarity between $\bold{Q}$ and $\bold{K}$, thereby resulting in deviation from the true attention matrix. To maintain the fidelity to the true similarity calculation by using binary $\bold{Q}$ and $\bold{K}$, we propose a frequency-based method to enhance similarity calculation. The core idea is illustrated in Figure~\ref{attentionmap} which involves using a simple non-subsampled Haar wavelet decomposition \textbf{(Only a 2 $\times$ 2 binary filter with negligible computational overhead)} to extract the local low-frequency ($\bold{X}^{L}$) and high-frequency ($\bold{X}^{H}$) components from the input ($\bold{X}$) of the attention module. The high-fidelity similarity $\bold{S}_{HF}$ is then calculated from the binarized enhanced $\bold{Q}_{e}$ and $\bold{K}_{e}$ obtained by concatenating different frequency components with the input. This process can be summarized as 
\begin{equation}
\label{similarity}
\begin{aligned}
\mathrm{\bold{S}_{HF}}=B(\bold{Q}_e)\otimes B(\bold{K}_e)^{T}
\end{aligned}
\end{equation}
where $\mathrm{B}()$ is the binary function defined in Eq.~\ref{eq2-1} while $\bold{Q}_{e}$ and $\bold{K}_{e}$ are obtained by
\begin{equation}
\label{HLqk2}
\begin{aligned}
\bold{Q}_e&=cat\left( \mathrm{BL}_{Q}^{L}\left( \bold{X}^{L} \right) ,\mathrm{BL}_{Q}^{H}\left( \bold{X}^{H} \right) \right)+\bold{X},\\
\bold{K}_e&=cat\left( \mathrm{BL}_{K}^{L}\left( \bold{X}^{L} \right) ,\mathrm{BL}_{K}^{H}\left( \bold{X}^{H} \right) \right)+\bold{X}.
\end{aligned}
\end{equation}
Here, $\mathrm{BL}_{Q}^{H}$ and $\mathrm{BL}_{Q}^{L}$ are binary linear layers that halve the output channel numbers compared with the binary linear layer for obtaining $\bold{V}$. 

From Eq.\ref{HLqk2}, we can see that our calculation leverages information from both the input and frequency-specific components to calculate similarity. Combining similarities derived from different frequencies helps mitigate biases introduced by binarization in any single frequency band. With the same idea of image matching techniques~\cite{ma2021image} that the more feature point data utilized for image matching, the superior the matching outcome achieved. As for the similarity calculation, applying this approach that superimposes the similarity results of each frequency component effectively enhances the similarity precision.

\subsection{Improved RPReLU}
The activation function plays a critical role in enhancing the performance of the binarized model. As shown in Eq.~\ref{act1}, compared with the PReLU~\cite{nair2010rectified} activation function,
the popular RPReLU~\cite{liu2020reactnet} is an effective activation function for binarization by shifting the feature distribution with learnable parameters $\boldsymbol{m}$ and $\boldsymbol{n}$. 
\begin{equation}
\label{act1}
\begin{aligned}[t]
\bold{PRELU}:& \bold{F}_{i,j}=\begin{cases}
	\bold{X}_{i,j}&		\bold{X}_{i,j}>0\\
	\boldsymbol{k}_i\cdot \bold{X}_{i,j}&		\bold{X}_{i,j}<0\\
\end{cases},
\\
\bold{RPRELU}:& \bold{F}_{i,j}=\begin{cases}
	\left( \bold{X}_{i,j}-\boldsymbol{m}_i \right) +\boldsymbol{n}_i&		\bold{X}_{i,j}>\boldsymbol{m}_i\\
	\boldsymbol{k}_i\left( \bold{X}_{i,j}-\boldsymbol{m}_i \right) +\boldsymbol{n}_i&		\bold{X}_{i,j}<\boldsymbol{m}_i\\
\end{cases},
\end{aligned}
\end{equation}
where $\bold{X}_{i,j}$ and $\bold{F}_{i,j}$ is an element of input $\bold{X}\in \mathbb{R} ^{C\times N}$ and activation $\bold{F}\in \mathbb{R} ^{C\times N}$, respectively. $\boldsymbol{m}\in \mathbb{R} ^{C}$ and $\boldsymbol{n}\in \mathbb{R} ^{C}$ are two learnable vectors that differ in each channel to shift the distribution of the feature. $\boldsymbol{k}\in \mathbb{R} ^{C}$ is the scale factor for the negative value.

However, for each channel of input (Each channel produces relatively independent features that can be analyzed separately), the effect of RPReLU on all tokens is the same. It only shifts the distribution of all values, which limits its capacity. We argue that modifying the overall distribution of the entire feature vector can further enhance its representational capacity. To achieve this, we propose an improved RPReLU, formulated as,
\begin{equation}
\label{act}
\begin{aligned}[t]
\bold{F}_{i,j}=\begin{cases}
	\left( \bold{X}_{i,j}-\boldsymbol{m}_{i} \right) +\boldsymbol{n}_{i}+\boldsymbol{t}_{j}&		\bold{X}_{i,j}>\boldsymbol{m}_{i}\\
	\boldsymbol{k}_i\left( \bold{X}_{i,j}-\boldsymbol{m}_{i} \right) +\boldsymbol{n}_{i}+\boldsymbol{t}_{j}&		\bold{X}_{i,j}<\boldsymbol{m}_{i}\\
\end{cases},
\end{aligned}
\end{equation}
where $\boldsymbol{t}\in \mathbb{R} ^{N}$ is different in each token for endowing the activation function with the ability to alter the overall distribution collaborating with $\boldsymbol{m}$ and $\boldsymbol{n}$. 

As shown in Eq.~\ref{act}, compared to the RPReLU, our improved RPReLU outputs distinct activation values for different tokens, achieving the reconstruction of multi-channel and multi-token feature distributions. This improvement can effectively expand the informativeness of the binary model. Notably, the parameters introduced by the proposed activation function are only $N+3C$ (due to $\boldsymbol{m},\boldsymbol{n},\boldsymbol{k}, and~\boldsymbol{t}$) for each activation layer, similar with FLOPs, which is negligible compared to the total parameters. 

\section{Experiment}
\label{exp}
In this section, we evaluate our method on both classification and segmentation tasks. Our method is implemented in PyTorch and all experiments are executed on a system with two NVIDIA A100 GPUs.

\noindent\textbf{Training Details.} 
Our method is trained using the AdamW optimizer with cosine annealing learning-rate decay, starting with an initial learning rate of $5\times 10^{-4}$ for all experiments. The training batch size is set to 256 for each classification dataset, with 150 epochs for the ImageNet-1K dataset and 300 epochs for the CIFAR-100 and Tiny-ImageNet datasets. For image and road segmentation tasks, the training batch sizes are 18 and 4, and the epochs are 50 and 100, respectively.

\noindent\textbf{Loss function.}~ For the classification task, the probabilistic distributions predicted by the teacher model provide richer information than one-hot labels, motivating us to leverage distillation learning with these soft labels to enhance our binary model. We choose the DeiT-small~\cite{touvron2021training} model as the teacher model due to its structural similarity to the student model. The overall loss function $L$ is:
\begin{equation}
\label{eqloss}
\begin{aligned}
L&=\left( 1-\lambda \right) L_{cls}+\lambda L_{dis}, 
\end{aligned}
\end{equation}
where $L_{cls}$ represents the cross-entropy loss between the model prediction and the one hot label, and $L_{dis}$ is the distillation loss between the teacher and student network outputs. $\lambda$ is set to 0.9. For the segmentation task, we only apply the cross-entropy loss between the model output and the label to optimize our binary model.\\
\noindent\textbf{Baseline Model.} We design a binary baseline model based on the ViT architecture, integrating several existing binarization techniques to ensure robust performance.\\
\indent $\bullet$ Incorporating RPReLU activation function and the RSign binarization function, as proposed by ReActNet~\cite{liu2020reactnet}.\\
\indent $\bullet$ Removing the class token from the ViT architecture and obtaining the final feature representation through a global pooling layer. \\
\indent $\bullet$ Implementing layer-wise residual connections as introduced in BiReal-Net~\cite{liu2018bi} to improve feature propagation.\\
\indent $\bullet$ Binarizing the activation, attention, and weights using the formulas defined in Eq.~\ref{eq2-1}, Eq.~\ref{eqatt}, and Eq.~\ref{eqw}
\subsection{Classification}
\begin{table}[htbp]
\setlength{\abovecaptionskip}{2pt}
\setlength{\belowcaptionskip}{0pt}
\renewcommand\arraystretch{1.0}
\caption{The comparison results for classification on CIFAR-100 and Tiny-ImageNet. W-A refers to the bit number of weights and activations for the corresponding method. The $\dagger$ and $\ddagger$ indicate the method applying the two-stage training~\cite{gao2024gsb} and decoupled training strategy~\cite{yin2024si}, respectively. All methods are training from scratch.}
		\label{tab2}
		\centering
  \setlength{\tabcolsep}{1.3pt}{
		\begin{tabular}{ccccc}  
			\Xhline{1.5pt}
			 Dataset   &Architecture                        & Method    & W-A       &  Top-1(\%)           \\
			\Xhline{1.0pt}	
            {\footnotesize\multirow{11}*{CIFAR-}}     &\multirow{4}*{Res-Net18~\cite{he2016deep}}                       & Real-value    & 32-32       &  72.5  \\ 
            {\footnotesize\multirow{11}*{100}}                              &  \multirow{4}*{(CNN)}                                             & Xnor-Net~\cite{rastegari2016xnor}    & 1-1       &  53.8  \\ 
             &                                               & IR-Net~\cite{qin2020forward}    & 1-1       &  64.5  \\  
                                          &                                               & RBNN~\cite{lin2020rotated}    & 1-1       &  65.3  \\ 
                                          &                                               & ReActNet~\cite{liu2020reactnet}    & 1-1       &  68.8  \\
                                          \cdashline{2-5}
                                          &\multirow{6}*{DeiT-Small~\cite{touvron2021training}}                      & Real-value    & 32-32       &  72.0  \\ 
                                          & \multirow{6}*{(ViT)}                                              & Q-ViT~\cite{li2022q}    & 2-2       &  68.3  \\ 
                                          &                                               & BiT~\cite{liu2022bit}    & 1-1       &  66.4  \\ 
                                          &                        &\textbf{Ours}                           & 1-1       &  \textbf{72.3}  \\ 
                                          \cdashline{3-5}
                                          &                                               & GSB$\dagger$~\cite{gao2024gsb}    & 1-1       &  71.1  \\ 
                                          &                        & BVT-IMA$\dagger$~\cite{wang2024bvt}    & 1-1       &  75.8  \\                  
                                          
                                          &                                             & \textbf{Ours$\dagger$}    & 1-1       &  \textbf{76.3}  \\

           \hline
            {\footnotesize\multirow{13}*{Tiny-}}  &\multirow{7}*{DeiT-Tiny~\cite{touvron2021training}}                       & Real-value    & 32-32       &  75.1    \\
            {\footnotesize\multirow{13}*{ImageNet}}                              & \multirow{7}*{(ViT)}                                              & BiT    & 1-1       &  24.3  \\
                                          &                                               & BiViT~\cite{he2023bivit}    & 1-1       &  37.5  \\
                                          &                                               & BFD~\cite{li2024bfd}    & 1-1       &  44.5  \\              
                                          &                                               & \textbf{Ours}    & 1-1       &  \textbf{46.6}  \\
                                          \cdashline{3-5}
                                          &                                               & Si-BiViT$\ddagger $~\cite{yin2024si}    & 1-1       &  49.8  \\
                                          &                                               & BVT-IMA$\dagger$    & 1-1       &  39.7  \\
                                          &                                               & \textbf{Ours$\dagger$}    & 1-1       &  \textbf{53.3}  \\
                                          \cdashline{2-5}
                                          &\multirow{6}*{DeiT-Small~\cite{touvron2021training}}                      & Real-value    & 32-32       &  78.6  \\
                                          & \multirow{6}*{(ViT)}                                              & BiT    & 1-1       &  38.8  \\
                                          &                                               & BiViT    & 1-1       &  44.7  \\
                                          &                                               & BFD    & 1-1       &  48.6  \\
                                          &                                               & \textbf{Ours}    & 1-1       &  \textbf{62.3}  \\
                                          \cdashline{3-5}
                                          &                                               & BVT-IMA$\dagger$    & 1-1       &  43.4  \\
                                          &                                               & \textbf{Ours$\dagger$}    & 1-1       &  \textbf{62.7}  \\
           
		\Xhline{1.5pt}
		\end{tabular}}
	\end{table}
\noindent \textbf{CIFAR-100~\cite{krizhevsky2009learning} and Tiny-ImageNet~\cite{wu2017tiny}} are two relatively small datasets commonly used for benchmarking binary ViTs. 
To demonstrate the effectiveness of our approach, we evaluate its performance using both CNN-based ResNet-18~\cite{he2016deep} and ViT-based DeiT~\cite{touvron2021training} as backbones. For a fair comparison, we categorize the compared methods into two main sub-groups based on their training settings: single-stage (default) and two-stage ($\dagger$).
All methods are trained from scratch, and the results are shown in Table~\ref{tab2}. Our method achieves superior performance for both datasets and all architectures. Specifically, under the single training setting, our method outperforms the state-of-the-art (SOTA) binarized ViT model, BiT, by \textbf{5.9\%} on the CIFAR-100 dataset. For the Tiny-ImageNet dataset, it surpasses BFD by \textbf{2.1\%} and \textbf{7.0\%} when using the DeiT-Tiny and DeiT-Small backbones, respectively. Under the two-stage training setting, which generally boosts performance compared to single-stage training, our method exceeds the SOTA method BVT-IMA by 0.5\% on CIFAR-100. For Tiny-ImageNet, it outperforms BVT-IMA by 13.6\% and 19.3\% with the DeiT-Tiny and DeiT-Small backbones, respectively. Additionally, it surpasses Si-BiViT, which employs a decoupled training strategy, by 3.5\%. Notably, our algorithm can even outperform the real-valued ViT and binary CNN models. For instance, under the two-stage training strategy, it achieves 76.3\% accuracy on the CIFAR-100 dataset, surpassing the real-valued DeiT-Small by \textbf{3.3\%} and the well-known binary CNN method ReActNet by \textbf{7.5\%}.

\begin{table}[tbh]
\renewcommand\arraystretch{1.0}
    	\caption{Results On ImageNet-1K. The $\dagger$ and $\ddagger$indicate the method applying the two-stage training~\cite{wang2024bvt} and decoupled training strategy~\cite{yin2024si}, respectively. $\mathrm{OPs}$ is defined as $\mathrm{OPs}=\frac{\mathrm{BOPs}}{64}+\mathrm{FLOPs}$~\cite{liu2020reactnet}. BOPs and FLOPs mean binary and float operations, respectively.}
		\label{tabimgnet}
		\centering
            \setlength{\tabcolsep}{1pt}{
		\begin{tabular}{ccccc}  
		\Xhline{1.3pt}
		\multirow{2}*{Network}     &\multirow{2}*{Methods}           & Size        &  OPs &  Top-1               \\
        		  &         & (MB)        &  ($\times 10^8$) &   (\%)                \\
		\Xhline{1.0pt}
            \multirow{6}*{DeiT-Tiny~\cite{touvron2021training}}	 &Real-valued       &  22.8 &   12.3      &  72.2                \\
            \cdashline{2-5}
            ~	 &BiT~\cite{liu2022bit}                   & \multirow{5}*{1.0}         &  \multirow{5}*{0.6}   &  21.7               \\
            ~	 &Bi-ViT~\cite{li2024bi}             &            &   &  28.7                 \\
            ~	 &BiBERT~\cite{qin2022bibert}             &           &    &  5.9                  \\
            ~	 &BVT-IMA$\dagger$~\cite{wang2024bvt}           &             &     &  30.0              \\
             ~	 &\textbf{Ours}      &             &    & \textbf{39.3$^{\textbf{(+9.3)}}$ }             \\
            \hline      
            \multirow{9}*{DeiT-Small~\cite{touvron2021training}}	 &Real-valued       & 88.4         &  45.5  &  79.9                \\
            \cdashline{2-5}
            ~	 &BiT            & \multirow{8}*{3.4}         &   \multirow{8}*{1.5}  &  30.7                 \\
            ~	 &BiViT~\cite{he2023bivit}            &         &    &  34.0                \\
            ~	 &BVT-IMA$\dagger$           &         &   &  48.0                \\
            ~	 &Bi-ViT            &         &    &  40.9                \\
            ~	 &BFD~\cite{li2024bfd}           &         &    &  47.5              \\

            ~	 &Si-BiViT$\ddagger$~\cite{yin2024si}           &         &    &  55.7             \\
                        ~	 &Baseline           &         &    &  49.5             \\
             ~	 &\textbf{Ours}      &             &    & \textbf{60.7$^{\textbf{(+5.0)}}$}             \\
                         \hline
            \multirow{6}*{Swin-Tiny~\cite{liu2021swin}}	 &Real-valued          & 114.2       &  44.9  &  81.2                \\
            \cdashline{2-5}
            ~	 &BiBERT         & \multirow{5}*{4.2}        & \multirow{5}*{1.5}  &  34.0                \\
            ~	 &RBONN~\cite{xu2022recurrent}         &         &   &  33.8                \\
            ~	 &Bi-Real Net~\cite{xu2022recurrent}         &         &   &  34.1                \\
            ~	 &Bi-ViT             &         &    &  55.5                 \\
            ~	 &\textbf{Ours}      &    &             & \textbf{61.5$^{\textbf{(+6.0)}}$}             \\
            \hline
              \multirow{3}*{BinaryViT~\cite{le2023binaryvit}}	 &Real-valued       & 90.4         &  46.0  &  79.9                \\
            \cdashline{2-5}
            ~	 &BinaryViT               & \multirow{2}*{3.5}         &   \multirow{2}*{0.79}  &  67.7                \\
            ~	 &\textbf{Ours}      &             &    & \textbf{68.4$^{\textbf{(+0.7)}}$ }             \\

              \hline

		\Xhline{1.5pt}
		\end{tabular}}
	\end{table}
\noindent \textbf{ImageNet-1K~\cite{deng2009imagenet}} contains approximately 1.2 million training images and 50,000 test images across 1,000 classes. The million-level data volume and the increased number of categories pose a significant challenge for the model. In this experiment, we test the performance on four variants architecture of ViT (DeiT-tiny, DeiT-small~\cite{touvron2021training}, BinaryViT~\cite{le2023binaryvit}, and Swin-Tiny~\cite{liu2021swin}). The results are shown in Table~\ref{tabimgnet}. Once again, we achieve superior performance by surpassing SOTA binarized ViTs across all four architectures by a large margin verifying the effectiveness of the proposed method. Our method outperforms the SOTA methods BVT-IMA by 9.3\% for the DeiT-Tiny network and outperforms the SOTA method Si-BiViT by 5.0\% for the DeiT-Small. 
Thanks to the integrated local attention and the multi-scale pyramid structure, our method also demonstrates excellent performance on BinaryViT and Swin-Tiny, achieving improvements of 0.7\% and 6.0\%, respectively. Compared to full-precision networks, our method achieves competitive performance while significantly reducing storage requirements and computational complexity, highlighting its efficiency and practicality.

\subsection{Image Segmentation} 
\noindent \textbf{ADE20K~\cite{zhou2019semantic}} is a challenging dataset including more than 20000 images with 150 categories with a limited amount of training data per class. We test each method by applying two evaluation metrics, including pixel accuracy (pixAcc) and mean Intersection-over-Union (mIoU). The results are shown in Table~\ref{tabade}, which demonstrates that our method achieves SOTA performance among current binary segmentation algorithms including both binary neural networks (BNNs) and binarized Vision Transformers (ViTs). It is worth noting that the BiDense~\cite{yin2024bidense}  applies channel-wise scaling factors to both activations and weights, which disrupts the acceleration process of the binary 1$\times$1 convolution operator and further reduces the practicality of the BiDense algorithm. Nevertheless, the pixAcc performance of our method with PVT-like architecture~\cite{wang2021pyramid} still surpasses that of BiDense, further highlighting its effectiveness.
\begin{table}[tb]
\setlength{\abovecaptionskip}{0pt}
\setlength{\belowcaptionskip}{0pt}
\setlength\tabcolsep{4pt}
\renewcommand\arraystretch{1.0}
\caption{Comparison results of image segmentation on ADE20K.}
\label{tabade}
\centering
		\begin{tabular}{ccccc}  
			\Xhline{1.5pt}
			   Method & Bit  & OPs (G)& pixAcc (\%) & mIoU  (\%) \\
			\Xhline{1.0pt}	
			  BNN~\cite{hubara2016binarized} & 1 & 4.84 & 61.69 & 8.68\\    
            \cdashline{1-5}
              ReActNet~\cite{liu2020reactnet} & 1 & 4.98 & 62.77 & 9.22\\     
              \cdashline{1-5}       
              AdaBin~\cite{tu2022adabin} & 1 & 5.24 & 59.47& 7.16\\   
             \cdashline{1-5}
			 BiSRNet~\cite{cai2024binarized} & 1 & 5.07 & 62.85 & 9.74\\  
             \cdashline{1-5}
             BiDense~\cite{yin2024bidense} & 1 & 5.37 & 67.25 & \textbf{18.75}\\  
             \hline
             BinaryViT~\cite{le2023binaryvit} & 1 & 4.95 & 67.60 & 17.25\\ 
             \cdashline{1-5}
             \textbf{BinaryViT} & \multirow{2}*{1} & \multirow{2}*{4.96} & \multirow{2}*{\textbf{68.72}} & \multirow{2}*{18.1}\\ 
             \textbf{+Ours} &  &  &  & \\
		\Xhline{1.5pt}
		\end{tabular}
\end{table} 
\begin{table}[tb]
\setlength{\abovecaptionskip}{2pt}
\setlength{\belowcaptionskip}{0pt}
\renewcommand\arraystretch{1.0}
\centering
\caption{Comparison results of road segmentation in aerial view.}
\setlength{\tabcolsep}{10pt}{
\begin{tabular}{cccc}
\Xhline{1.5pt}
Encoder   & W-A  & mAcc & mIou   \\
\Xhline{1.0pt}
ResNet-34~\cite{he2016deep}  &32-32  & 85.4 & 77.8      \\
ReActNet~\cite{liu2020reactnet}  &1-1  & 76.5 & 63.6      \\
\cdashline{1-4} 
DeiT-Small+Baseline  &1-1  & 85.9 & 72.5      \\
\textbf{DeiT-Small+Ours}  &1-1  & \textbf{91.1} &\textbf{ 76.5 }    \\
\cdashline{1-4} 
BinaryViT~\cite{le2023binaryvit}  &1-1  & 90.9 & 82.9      \\
\textbf{BinaryViT+Ours}  &1-1  & \textbf{91.2} & \textbf{83.6}     \\
\Xhline{1.5pt}
\end{tabular}}
\label{Tabrslvf} 
\end{table}

\noindent \textbf{Road Segmentation} is another crucial task in computer vision, particularly for applications such as autonomous driving and urban planning. We evaluate the performance of our method for road segmentation using the BEV perspective. We employ the RS-LVF dataset~\cite{yin2024pathfinder}, which includes 1,000 aerial images, point clouds (typically projected into the image coordinate system using a calibration file), and corresponding road labels. This dataset is specifically designed for per-pixel binary classification tasks in road segmentation, where the data inherently exhibits a severe class imbalance that the model must address. We assess performance using the Mean Intersection over Union (mIoU) and mean accuracy metrics. The results are presented in Table~\ref{Tabrslvf}. Each method is based on an encoder-decoder architecture, similar to U-Net~\cite{ronneberger2015u}, with the decoder module utilizing a ResNet structure that progressively increases resolution. Notably, different from ReActNet~\cite{liu2020reactnet} and our method, the ResNet-34 baseline maintains full precision for both weights and activations.  Compared to the DeiT-Small architecture, the BinaryViT architecture incorporates a multi-scale pyramid structure, making it more suitable for pixel-level segmentation tasks. As shown in Table~\ref{Tabrslvf}, our proposed DIDB-ViT model achieves higher segmentation accuracy than the baseline, the original BinaryViT, and the CNN-based ReActNet, even outperforming full-precision methods.
\subsection{Ablation Study}
\noindent\textbf{Impact of Each Proposed Module.} Our method comprises three modules: Differential-Informative Binary Attention (\textbf{DIBA}), High-Fidelity Similarity Calculation (\textbf{HFSC}), and Improved RPReLU (\textbf{IRPReLU}) activation function. As demonstrated in Table~\ref{tabxr}, the performance variations of the proposed binary method using the DeiT-small model on ImageNet-1k are shown by sequentially removing each proposed module, thus validating the contribution of each component. From the results, we can clearly see that DIBA and HFSC are critical modules improving the accuracies by 5.2\% and 4.2\% respectively. This verifies the importance of incorporating differential information and frequency-based similarity calculation. IRPReLU is also an imperative module as it can further improve the accuracy by 1.8\%.

In Table~\ref{tabxr2}, we observe that incorporating the HFSC module leads to superior performance compared to models that omit it, particularly under the two-stage (\textbf{TS}) training strategy. This highlights the ability of the HFSC module to enhance model optimization, showcasing its effectiveness in improving the model's learning process. The introduction of two distinct gradient pathways (one for high-frequency and one for low-frequency information) further supports the model’s ability to better activate each element, thus improving the optimization efficiency of binary models. This reinforces the positive impact of the HFSC module in enhancing model performance, especially when combined with the two-stage training technique.
\begin{table}[htbp]
\setlength{\abovecaptionskip}{2pt}
\setlength{\belowcaptionskip}{2pt}
\setlength\tabcolsep{7pt}
\renewcommand\arraystretch{1.0}
\caption{Ablation study for the proposed modules with DeiT-small model on the ImageNet-1k.}
\label{tabxr}
\centering
		\begin{tabular}{ccccc}  
			\Xhline{1.3pt}
			   Baseline & DIBA & HFSC & IRPReLU & Top1 (\%) \\
			\Xhline{1.0pt}	
			
              $\checkmark$&&&&49.5 \\
            \cdashline{1-5}
            $\checkmark$&$\checkmark$&&&54.7  \\ 
                 \cdashline{1-5}       
               $\checkmark$&$\checkmark$&$\checkmark$&& 58.9 \\  
              \cdashline{1-5}
			    $\checkmark$&$\checkmark$&$\checkmark$&$\checkmark$& 60.7\\    
		\Xhline{1.5pt}
		\end{tabular}
\end{table} 

\begin{table}[htbp]
\setlength{\abovecaptionskip}{0pt}
\setlength{\belowcaptionskip}{-2pt}
\setlength\tabcolsep{30pt}
\renewcommand\arraystretch{1.0}
\caption{Ablation study for the proposed binary method with DeiT-small model on the ImageNet-1k.}
\label{tabxr2}
\centering
		\begin{tabular}{lc}  
			\Xhline{1.5pt}
			   Method  & Top1 (\%) \\
			\Xhline{1.0pt}	
                  Baseline  & 49.5 \\
                  \cdashline{1-2}
			     + DIBA    &      54.7     \\
                  + DIBA,~~~TS    &     58.3      \\
                  + DIBA,~~~HFSC    &    58.9       \\
  
		\Xhline{1.5pt}
		\end{tabular}
\end{table}
\noindent\textbf{Impact of coefficient $\lambda$.} Based on the CIFAR-100 dataset, we examine the influence introduced by the balance parameters $\lambda$ on the classification performance of the DIDB-ViT model. The results are shown in Table~\ref{tablamda}, from which we can conclude that the model achieves the highest accuracy when the hyperparameter $\lambda$ is set to 0.9. 
\begin{table}[tb]
\setlength\tabcolsep{1pt}
\setlength{\abovecaptionskip}{2pt}
\setlength{\belowcaptionskip}{-5pt}
\renewcommand\arraystretch{1.3}
    	\caption{Ablation study for DIDB-ViT on the CIFAR-100 dataset with different hyperparameters $\lambda$. }
		\label{tablamda}
		\centering
		\begin{tabular}{cccccccccccc}  
			\Xhline{1.5pt}
			  $\lambda$&  0&  0.1  & 0.2 & 0.3 & 0.4 & 0.5 & 0.6 &0.7& 0.8&0.9&1.0 \\
			\Xhline{1.0pt}	
			  Top1 &68.7 & 69.6   & 70.2 &  70.5 & 70.8  & 71.3 &  71.4 & 71.7 & 72.1 &72.3 &72.1\\    
		\Xhline{1.5pt}
		\end{tabular}
	\end{table}
    
\noindent\textbf{Impact of Receptive Field $\theta$ in DIBA .} Based on the ImageNet-1K dataset, we examine the influence introduced by the Receptive Field $\theta$ of additional group binary convolution layer in DIBA  on the classification performance of the DIDB-ViT model. The result is shown in Tab.~\ref{tabtheta}, from which we can conclude that the model achieves the balance between performance and computational effort when the Receptive Field $\theta$ is set to $3 \times 3$. 
\begin{table}[htbp]
\setlength\tabcolsep{13pt}
\setlength{\abovecaptionskip}{0pt}
\setlength{\belowcaptionskip}{0pt}
\renewcommand\arraystretch{1.0}
    	\caption{Ablation study for DIDB-ViT with different Receptive Field $\theta$ in DIBA. }
		\label{tabtheta}
		\centering
		\begin{tabular}{ccc}  
			\Xhline{1.5pt}
			  Receptive Field $\theta$& OPs ($\times 10^6$) &  Top1 (\%)   \\
			\Xhline{1.0pt}	
			  $3 \times 3$ & 0.33& 60.7   \\  
                $5 \times 5$ & 0.92& 60.8   \\ 
                $7 \times 7$ & 1.81& 60.2  \\ 
		\Xhline{1.5pt}
		\end{tabular}
	\end{table}
    \vspace{-0.4cm}
\section{Conclusion}
This paper introduces DIDB-ViT, a high-fidelity differential information driven binary vision transformer designed to narrow the performance gap between binary ViTs and their full-precision counterparts while preserving the original ViT architecture. Our approach includes a differential-informative attention module to restore lost differential information and improve token interaction. We further enhance the accuracy of similarity calculations between binary $Q$ and $K$ tensors using a non-subsampled discrete Haar wavelet transform to retain both high- and low-frequency components. To expand the representational capacity of the binary model, we propose an improved activation function inspired by RPReLU, enabling more expressive feature distributions. Extensive experiments on classification and segmentation datasets demonstrate the superior performance of our method, highlighting its effectiveness for resource-constrained environments.

\noindent \textbf{Acknowledgements.} This work was supported by the Fundo para o Desenvolvimento das Ciências e da Tecnologia of Macau (FDCT) with Reference No. 0067/2023/AFJ, No. 0117/2024/RIB2

{\small
\bibliographystyle{ieee_fullname}
\bibliography{egbib}
}

\newpage

\maketitlesupplementary
\setcounter{table}{0}
\setcounter{figure}{0}
\setcounter{section}{0}
\setcounter{equation}{0}
\setcounter{Proposition}{0}

\section{The Gradient Backpropagation of High-Fidelity Similarity Calculation}
For instance, consider the gradient between the output $\bold{O}$ and
the input $\bold{X}$ in the attention module. Let the element of $\bold{O}$ be $\bold{O}^{k,l}$ and the element of $\bold{X}$ be $\bold{X}^{m,n}$, we have
\begin{equation}
\label{eqB1}
\begin{aligned}[t]
\frac{\partial \bold{O}^{k,l}}{\partial \bold{X}^{m,n}}=\partial \bold{V}+\partial \bold{Q}+\partial \bold{K}.
\\
\end{aligned}
\end{equation}
\begin{equation}
\label{eqB2}
\begin{aligned}[t]
&\partial \bold{V}=\sum_{i=1}^t{\frac{\partial \bold{O}^{k,l}}{\partial B\left( \bold{V}^{i,l} \right)}}\cdot \frac{\partial B\left( \bold{V}^{i,l} \right)}{\partial \bold{V}^{i,l}}\cdot \frac{\partial \bold{V}^{i,l}}{\partial \bold{X}^{m,n}},
\\
&\frac{\partial \bold{O}^{k,l}}{\partial B\left( \bold{V}^{i,l} \right)}=\bold{A}_{as}^{k,i},
\frac{\partial B\left( \bold{V}^{i,l} \right)}{\partial \bold{V}^{i,l}}=\left\{ \begin{matrix}
	1&		\left| \bold{V}^{i,l} \right|\leqslant 1\\
	0&		others\\
\end{matrix} \right. 
\\
&\frac{\partial \bold{V}^{i,l}}{\partial \bold{X}^{m,n}}=\left\{ \begin{matrix}
	B(\bold{W}_{V}^{i,m})&		\left( \begin{array}{c}
	n=l,\\
	\left| \left( \bold{X}^{m,n} \right) \right|\leqslant 1\\
\end{array} \right)\\
	0&		others\\
\end{matrix} \right. ,
\end{aligned}
\end{equation}
where $A_{as}^{k,:} \in \mathbb{R} ^{t}$ and $A_{bs}^{k,:} \in \mathbb{R} ^{t}$ are the attention matrix after and before the scaling process ($softmax$, and $\sqrt{c}$) respectively. $\bold{W}_{V}^{i,m}$ is the weight of linear layer for obtaining $\bold{V}$.  $k\& m\in \left[ 1,t \right] ,n\& l\in \left[ 1,c \right]$. $t$ means the token number and $c$ is the channel number. We omit the batch size and head dimension for simplicity for each activation. $B()$ means binarization function. $B\left(\bold{X}\right)$ is the binary input tensor of the attention module. 
\begin{equation}
\label{eq8}
\begin{aligned}[t]
\partial Q&=\frac{\partial O^{k,l}}{\partial B\left( A_{as}^{k,:} \right)}\cdot \frac{\partial B\left( A_{as}^{k,:} \right)}{\partial A_{as}^{k,:}}\cdot \frac{\partial A_{as}^{k,:}}{\partial A_{bs}^{k,:}}\\
&\cdot \frac{\partial A_{bs}^{k,:}}{\partial B\left( Q^{k,:} \right)}\cdot \frac{\partial B\left( Q^{k,:} \right)}{\partial Q^{k,:}}\cdot \frac{\partial Q^{k,:}}{\partial X^{m,n}}
\\
\frac{\partial O^{k,l}}{\partial B\left( A_{as}^{k,:} \right)}&=B\left( V^{:,l} \right) \,\,,\\
\frac{\partial B\left( A_{as}^{k,:} \right)}{\partial A_{as}^{k,:}}&=1_{0.5\leqslant A^{k,:}\leqslant 1}, \\
\frac{\partial A_{as}^{k,:}}{\partial A_{bs}^{k,:}}&=\frac{A^{k,:}\otimes \left( 1-A^{k,:} \right)}{\sqrt{c}},\\
\frac{\partial A_{bs}^{k,:}}{\partial B\left( Q^{k,:} \right)}&=\sum_{i=1}^t{B\left( (K^{i,:}) \right)},\\
\frac{\partial B\left( Q^{k,:} \right)}{\partial Q^{k,:}}&=1_{\left| Q^{k,:} \right|\leqslant 1},\\
\frac{\partial Q^{k,:}}{\partial X^{m,n}}&=\left\{ \begin{matrix}
	\sum_{i=1}^c{B\left( W_{q}^{n,i} \right)}&		\left( \begin{array}{c}
	k=m,\\
	\left| X^{m,n} \right|\leqslant 1\\
\end{array} \right)\\
	0&		others\\
\end{matrix} \right. ,  
\end{aligned}
\end{equation}
Where $\otimes$ denotes Hadamard product.
\begin{equation}
\label{eq82}
\begin{aligned}[t]
\partial K&=\frac{\partial O^{k,l}}{\partial B\left( A_{as}^{k,m} \right)}\cdot \frac{\partial B\left( A_{as}^{k,m} \right)}{\partial A_{as}^{k,m}}\cdot \frac{\partial A_{as}^{k,m}}{\partial A_{bs}^{k,m}}\cdot\\
&\frac{\partial A_{bs}^{k,m}}{\partial B\left( (K^{m,:})^T \right)}\cdot \frac{\partial B\left( (K^{m,:})^T \right)}{\partial (K^{m,:})^T}\cdot \frac{\partial (K^{m,:})^T}{\partial X^{m,n}},
\\
\frac{\partial O^{k,l}}{\partial B\left( A_{as}^{k,m} \right)}&=B\left( V^{m,l} \right) ,\\
\frac{\partial B\left( A_{as}^{k,m} \right)}{\partial A_{as}^{k,m}}&=\mathbf{1}_{0.5\leqslant A^{k,m}\leqslant 1},\\
\frac{\partial A_{as}^{k,m}}{\partial A_{bs}^{k,m}}&=\frac{A^{k,m}\otimes \left( 1-A^{k,m} \right)}{\sqrt{c}}
\\
\frac{\partial A_{bs}^{k,m}}{\partial B\left( (K^{m,:})^T \right)}&=B\left( Q^{k,:} \right) ,\\
\frac{\partial B\left( (K^{m,:})^T \right)}{\partial (K^{m,:})^T}&=\mathbf{1}_{\left| (K^{m,:})^T \right|\leqslant 1},\\
\frac{\partial (K^{m,:})^T}{\partial X^{m,n}}&=\left\{ \begin{matrix}
	\sum_{i=1}^c{B\left( W_{k}^{n,i} \right)}&		\left| X^{m,n} \right|\leqslant 1\\
	0&		others\\
\end{matrix} \right. .
\end{aligned}
\end{equation}

The stacking of binarization functions exacerbates the issue of gradient vanishing~\cite{gao2024gsb} (The gradient of input values greater than 1 or less than -1 will be set to zero.), which leads to incomplete optimization of the input tensor. To solve this problem, we propose
the high-fidelity similarity module $\bold{S}_{HF}$, which is as follows,
\begin{equation}
\label{similarity}
\begin{aligned}
\mathrm{\bold{S}_{HF}}=B(\bold{Q}_e)B(\bold{K}_e)^{T}
\end{aligned}
\end{equation}
where $\mathrm{B}()$ is the binary function while $\bold{Q}_{e}$ and $\bold{K}_{e}$ are obtained by
\begin{equation}
\label{HLqk2}
\begin{aligned}
\bold{Q}_e&=cat\left( \mathrm{BL}_{Q}^{L}\left( \bold{X}^{L} \right) ,\mathrm{BL}_{Q}^{H}\left( \bold{X}^{H} \right) \right)+\bold{X},\\
\bold{K}_e&=cat\left( \mathrm{BL}_{K}^{L}\left( \bold{X}^{L} \right) ,\mathrm{BL}_{K}^{H}\left( \bold{X}^{H} \right) \right)+\bold{X}.\\
\bold{X}_{i,j}^{L}&=\bold{X}_{i-1,j-1}+\bold{X}_{i-1,j+1}+\bold{X}_{i+1,j-1}+\bold{X}_{i+1,j+1},\\
\bold{X}_{i,j}^{H}&=\bold{X}_{i-1,j-1}+\bold{X}_{i+1,j+1}-\bold{X}_{i-1,j+1}-\bold{X}_{i+1,j-1},
\end{aligned}
\end{equation}
Here, $\mathrm{BL}_{Q}^{H}$ and $\mathrm{BL}_{Q}^{L}$ are binary linear layers that halve the output channel numbers compared with the binary linear layer for obtaining $\bold{V}$.

From Eq.~\ref{HLqk2}, we could find out that the additional information introduced by $\bold{S}_{HF}$ broadens the interaction and increases the number of pathways for the gradient from the output of the attention module to the input, improving the training process.
\section{The Improved RPReLU}
\begin{figure*}
\centering
\includegraphics[width=7.0in]{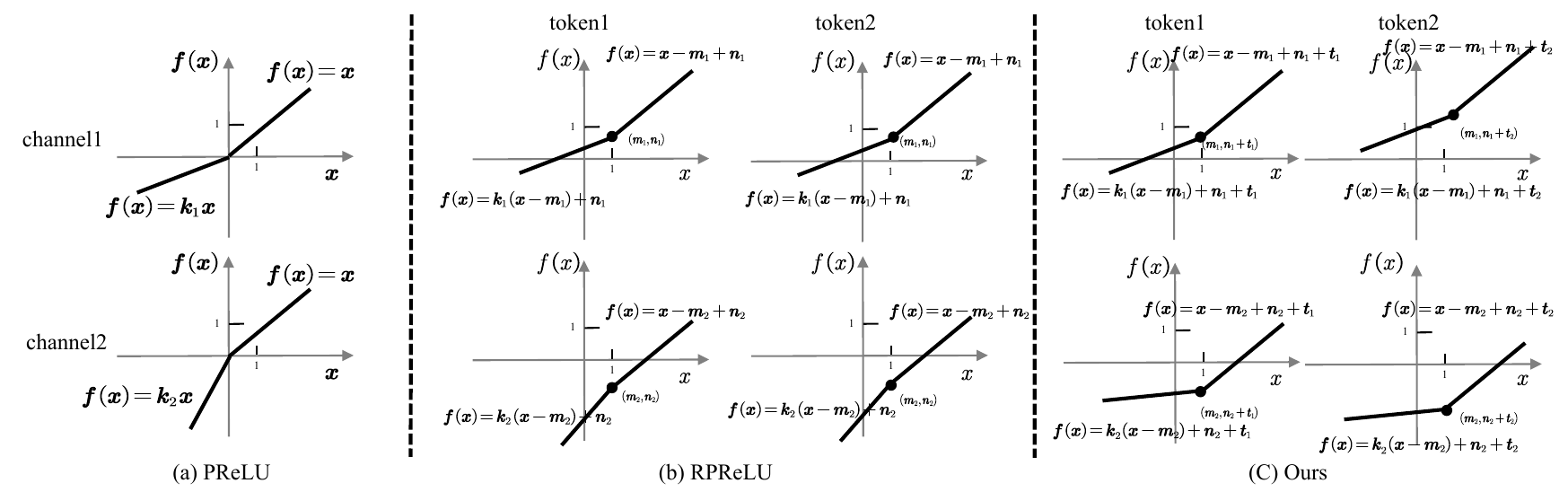}
\caption{PReLU and RPReLU versus our improved RPReLU. For an input vector $\boldsymbol{X}\in \mathbb{R} ^{2\times 2}$, our improved RPReLU assigns different activation values to different tokens, achieving reconstruction of feature distributions across multiple channels and tokens.}
\label{figactivation}
\end{figure*}
\begin{figure*}
\centering
\includegraphics[width=6.5in]{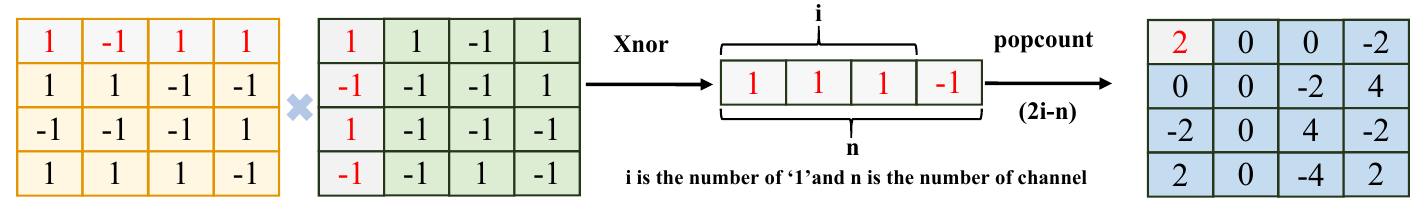}
\caption{The multiplication between binary vectors can be implemented by the Xnor and popcount. The result is $2i-n$}
\label{figxnor}
\end{figure*}
To achieve shifting the overall distribution of the entire feature vector, we propose an improved RPReLU, formulated as,
\begin{equation}
\label{act}
\begin{aligned}[t]
\bold{F}_{i,j}=\begin{cases}
	\left( \bold{X}_{i,j}-\boldsymbol{m}_{i} \right) +\boldsymbol{n}_{i}+\boldsymbol{t}_{j}&		\bold{X}_{i,j}>\boldsymbol{m}_{i}\\
	\boldsymbol{k}_i\left( \bold{X}_{i,j}-\boldsymbol{m}_{i} \right) +\boldsymbol{n}_{i}+\boldsymbol{t}_{j}&		\bold{X}_{i,j}<\boldsymbol{m}_{i}\\
\end{cases},
\end{aligned}
\end{equation}
As shown in Fig.~\ref{figactivation}, we
take an input vector $\bold{X}\in \mathbb{R} ^{2\times 2}$ as an example. For each channel of input (Each channel produces relatively independent features that can be analyzed separately), the effect of RPReLU on all tokens is the same. Compared to the RPReLU, our improved RPReLU outputs distinct activation values for different tokens, achieving the reconstruction of multi-channel and multi-token feature distributions. This improvement can effectively expand the informativeness of the binary model. The proof of this theory can be found in Observation~\ref{lemma1} \\
\begin{Proposition}
\label{lemma1}
 Adding a learnable parameter to each element of a matrix can alter the overall distribution of values, but adding the same learnable parameter to all elements of a matrix cannot change the overall distribution.
\end{Proposition}

\noindent \textbf{Detailed illustration.} To demonstrate the above Observation, we can analyze two scenarios: adding a unique learnable parameter to each element of the matrix and adding a single learnable parameter to all elements of the matrix.\\
\noindent \textbf{1. Adding a Unique Learnable Parameter to Each Element.}\\
Let $A\in \mathbb{R} ^{m\times n}$ be the original matrix. We add a unique learnable parameter matrix $P\in \mathbb{R} ^{m\times n}$ to $A$, where $p_{ij}$ is an independent learning parameter and $a_{ij}$ is the element of $A$. The updated matrix is,
\begin{equation}
\label{iact1}
\begin{aligned}[t]
A^{\prime}=A+P
\end{aligned}
\end{equation}
The mean and variance of the Original matrix $A$ are as follows, 
\begin{equation}
\label{iact2}
\begin{aligned}[t]
\mu _A&=\frac{1}{mn}\sum_{i=1}^m{\sum_{j=1}^n{a_{ij}}},\\
\sigma _{A}^{2}&=\frac{1}{mn}\sum_{i=1}^m{\sum_{j=1}^n{\left( a_{ij}-\mu _A \right) ^2}}.
\end{aligned}
\end{equation}
The mean and variance of the updated matrix $A^{\prime}$ are,
\begin{equation}
\label{iact3}
\begin{aligned}[t]
\mu _{A^{\prime}}=\frac{1}{mn}\sum_{i=1}^m{\sum_{j=1}^n{\left( a_{ij}+p_{ij} \right)}=\mu _A+\mu _P,}\\
\sigma _{A^{\prime}}^{2}=\frac{1}{mn}\sum_{i=1}^m{\sum_{j=1}^n{\left( \left( a_{ij}+p_{ij} \right) -\mu _{A^{\prime}} \right) ^2}},
\end{aligned}
\end{equation}
where $\mu _P=\frac{1}{mn}\sum_{i=1}^m{\sum_{j=1}^n{p_{ij}}}$. $a_{ij}^{\prime}=a_{ij}+p_{ij}$ and $a_{ij}^{\prime}$ is the element of $A^{\prime}$.
Due to $p_{ij}$ is a learnable parameter, during the learning process, the gradient of $p_{ij}$ is,
\begin{equation}
\label{iact4}
\begin{aligned}[t]
{\frac{\partial L}{\partial p_{ij}}}=\frac{\partial L}{\partial a_{ij}^{\prime}}.\frac{\partial a_{ij}^{\prime}}{\partial p_{ij}}=\frac{\partial L}{\partial a_{ij}^{\prime}},
\end{aligned}
\end{equation}
where $L$ is the Loss function. $P$ will be dynamically adjusted through the learning process, and $A^{\prime}$ will also be adjusted accordingly, thereby altering the distribution of $A^{\prime}$.\\
\noindent \textbf{2. Adding a Single Learnable Parameter to All Elements.}\\
In this situation, the learnable parameter becomes a scalar $p\in \mathbb{R} ^{1\times 1}$. The updated matrix is,

\begin{equation}
\label{iact5}
\begin{aligned}[t]
A''=A+p\cdot\mathbf{1},
\end{aligned}
\end{equation}
where $\mathbf{1}$ is a all-one matrix with the same shape as $A$.\\
The mean and variance of the updated matrix $A^{\prime}$ are,
\begin{equation}
\label{iact6}
\begin{aligned}[t]
\mu _{A''}&=\frac{1}{mn}\sum_{i=1}^m{\sum_{j=1}^n{\left( a_{ij}+p \right)}=\mu _A+p,}\\
\sigma _{A''}^{2}&=\frac{1}{mn}\sum_{i=1}^m{\sum_{j=1}^n{\left( \left( a_{ij}+p \right) -\mu _{A''} \right) ^2}},\\
&=\frac{1}{mn}\sum_{i=1}^m{\sum_{j=1}^n{\left( a_{ij} -\mu _{A} \right) ^2}}=\sigma _{A}^{2}.
\end{aligned}
\end{equation}
From this, it can be seen that adding the same learnable parameter $p$ will only change the mean of the matrix (shifting the distribution), without altering the variance or shape of the distribution of the matrix.\\
The Back Propagation of $p$ is as follows,
\begin{equation}
\label{iact7}
\begin{aligned}[t]
\frac{\partial L}{\partial p}=\sum_{i=1}^m{\sum_{j=1}^n{\frac{\partial L}{\partial a_{ij}^{''}}.\frac{\partial a_{ij}^{''}}{\partial p_{ij}}}}=\sum_{i=1}^m{\sum_{j=1}^n{\frac{\partial L}{\partial a_{ij}^{''}}}}.
\end{aligned}
\end{equation}
Since $p$ is a scalar, it can only make the same adjustment to all elements of the matrix, without changing the relative relationships between elements.

From the Observation~\ref{lemma1}, we can find out that for each channel of feature, RPReLU can only shift the distribution of the feature value but the proposed activation function can achieve the reconstruction of multi-channel and multi-token feature
distributions.

Then, we discuss the optimization problem of the introduced parameters, and the gradient back-propagation process for each introduced parameter is shown in Eq.~\ref{actb}.

\begin{equation}
\label{actb}
\begin{aligned}[t]
&\frac{\partial L}{\partial \boldsymbol{m}_{i}}=\sum_{j=1}^N{\frac{\partial L}{\partial \bold{F}_{i,j}}}\cdot \frac{\partial \left( \bold{X}_{i,j}-\boldsymbol{m}_{i} \right)}{\partial \boldsymbol{m}_{i}}\cdot \mathbb{I} _{i,j},
\\
& \mathbb{I} _{i,j}=\left\{ \begin{matrix}
	1&		\bold{X}_{i,j}\geqslant \boldsymbol{m}_i\\
	\boldsymbol{k}_i&		\bold{X}_{i,j}<\boldsymbol{m}_i\\
\end{matrix} \right. , \frac{\partial L}{\partial \boldsymbol{k}_i}=\sum_{j=1}^N{\frac{\partial L}{\partial \bold{F}_{i,j}}\cdot \frac{\partial \bold{F}_{i,j}}{\partial \boldsymbol{k}_i},}
\\
&\frac{\partial L}{\partial \boldsymbol{n}_{i}}=\sum_{j=1}^N{\frac{\partial L}{\partial \bold{F}_{i,j}}}\cdot \frac{\partial \bold{F}_{i,j}}{\partial \boldsymbol{n}_{i}},
\frac{\partial L}{\partial \boldsymbol{t}_{j}}=\sum_{i=1}^C{\frac{\partial L}{\partial \bold{F}_{i,j}}}\cdot \frac{\partial \bold{F}_{i,j}}{\partial \boldsymbol{t}_{j}},
\end{aligned}
\end{equation}
where ${\frac{\partial L}{\partial \bold{F}_{i,j}}}$ is the gradient between the loss function and the output of the proposed activation layer, which comes from the deeper layer. $\frac{\partial \left( \bold{X}_{i,j}-\boldsymbol{m}_{i} \right)}{\partial \boldsymbol{m}_{i}} = -1$, $\frac{\partial \bold{F}_{i,j}}{\partial \boldsymbol{n}_{i}}=1$, and $\frac{\partial \bold{F}_{i,j}}{\partial \boldsymbol{t}_{j}}=1$. if $\bold{X}_{i,j}<\boldsymbol{m}_{i}$, $\frac{\partial \bold{F}_{i,j}}{\partial \boldsymbol{k}_{i}} = 1$, else $\frac{\partial \bold{F}_{i,j}}{\partial \boldsymbol{k}_{i}} = 0$. 

\section{Xnor+Popcount}
During the inference process of the model, the primary concept of the model binarization technique involves converting each numerical value linked with matrix multiplication to either 1 or -1 and applying $Xnor$ and $popcount$ operations to replace the multiplication of binary vectors for extremely high computational efficiency (Fig.~\ref{figxnor}).
\section{Evaluation}
\noindent\textbf{The ablation study about the latency}
To obtain a latency result comparison between our method and the other binary ViT method, we first transfer the Pytorch code of each method to the ONNX version. Then, we utilize the BOLT toolbox~\cite{bolt} to implement each method to the edge device based on an ARM Cortex-A76 CPU (without CUDA). The result is shown in the Tab.~\ref{tablatency}. The image resolution of the test image is 224$\times$224. Due to the lack of optimization and deployment methods for the specific modules in the ViT structure, the acceleration results of binarized ViT cannot achieve an ideal acceleration state the same as the BNN. Therefore, further deployment techniques must be developed to show the full advantages of binary vision transformers on edge devices.
\begin{table}[htbp]
\renewcommand\arraystretch{1.1}
    	\caption{The latency result of the proposed method.}
		\label{tablatency}
		\centering
         \setlength{\tabcolsep}{16pt}{
		\begin{tabular}{ccc}  
			\Xhline{1pt}
			  Method & W/A (bit) & Latency (ms)  \\
			\Xhline{1pt}	
       Ours & 32/32 & 863 \\
       Baseline & 1/1 & 342 \\    
       Ours & 1/1 & 361 \\
		\Xhline{1pt}
		\end{tabular}}
	\end{table}

\end{document}